\documentclass{jdmdh}
\usepackage{array}
\usepackage{pgfplots}
\usepackage{float}
\titlespacing*{\section}
{0pt}{4.5ex plus 1ex minus .2ex}{2.5ex plus .2ex}
\titlespacing*{\subsection}
{0pt}{4.5ex plus 1ex minus .2ex}{2.5ex plus .2ex}
\titlespacing*{\subsubsection}
{0pt}{4.5ex plus 1ex minus .2ex}{2.5ex plus .2ex}
\title{Combining Morphological and Histogram based Text Line Segmentation in the OCR Context}
\author[1]{Pit Schneider}
\affil[1]{National Library of Luxembourg, Luxembourg} 

\corrauthor{Pit Schneider}{pit.schneider@bnl.etat.lu}

\begin{document}

\maketitle

\abstract{Text line segmentation is one of the pre-stages of modern optical character recognition systems. The algorithmic approach proposed by this paper has been designed for this exact purpose. Its main characteristic is the combination of two different techniques, morphological image operations and horizontal histogram projections. The method was developed to be applied on a historic data collection that commonly features quality issues, such as degraded paper, blurred text, or presence of noise. For that reason, the segmenter in question could be of particular interest for cultural institutions, that want access to robust line bounding boxes for a given historic document. Because of the promising segmentation results that are joined by low computational cost, the algorithm was incorporated into the OCR pipeline of the National Library of Luxembourg, in the context of the initiative of reprocessing their historic newspaper collection. The general contribution of this paper is to outline the approach and to evaluate the gains in terms of accuracy and speed, comparing it to the segmentation algorithm bundled with the used open source OCR software.}


\section{Context}
While adapting modern open source OCR software to reprocess a large collection of historic newspapers, ranging from years 1841 to 1954, the National Library of Luxembourg (BnL) explored ways to segment the newspaper scans into individual text lines. Pursuing this goal, a method was developed that integrates into a larger OCR pipeline by sitting just in between the binarization and font recognition processes. \\ \\
The used open source $kraken$ software by \citet{kraken}, forked from the OCRopus OCR System by \citet{ocropus}, that implements the character recognition step of the pipeline, already comes equipped with a segmentation method. That's in the form of the $kraken.pageseg.segment$ function that mostly relies on different filters from $scipy.ndimage$ (\citet{scipy}). This algorithm, as published with version 2.0.8 \footnote{Most recent release at the time of this work. During peer review, the newest version (3.0.4) leaves the $kraken.pageseg.segment$ functionality unchanged, but now also includes $kraken.blla.segment$ (model-based).} and in the following referred to as \textsc{Bench}, essentially served as a benchmark in the context of the development of an own solution, designed for the precise needs of BnL. \\ \\
The motivation for this initiative was threefold: \\
\begin{itemize}
    \item Due to the large volume of around 6 million newspaper articles, building a computationally efficient segmenter was the main priority and source of motivation for the project. Next to running faster than \textsc{Bench}, the aim was for the method to only take up a fraction of the time needed for the OCR pipeline's character recognition functionality. \\
    \item Given the historic data target, another goal was to be able to deal with the common properties of the data, such as degraded paper quality, blurred text, or presence of noise. Problems, as seen with example text block image depicted in Figure~\ref{fig:input}, repeatedly appear in the dataset.
    \begin{figure}[H]
    \centering
    \includegraphics[width=0.66\textwidth]{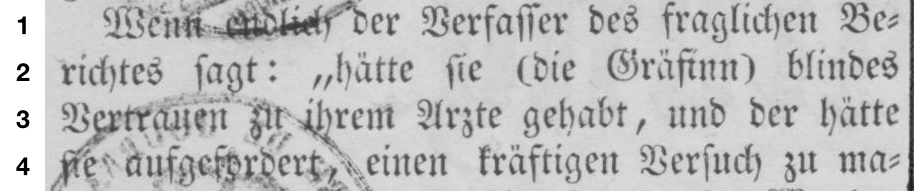}
    \captionsetup{format=hang, margin=0.0\textwidth}
    \caption{Example text block of historic newspaper collection, featuring four text lines and the potential issues of background noise (stamp) and border presence (right).}
    \label{fig:input}
    \end{figure}
    \item Furthermore, \textsc{Bench} is advertised as a page segmenter, including layout analysis. The algorithm can however be passed a parameter, indicating that a single column of text (i.e. zero column seperators) is to be expected. Naturally this was the chosen configuration for comparison purposes. Since the single column property is always valid for the pre-segmented text blocks in the BnL dataset, a solution, purely designed for line detection within a single block, was favored. The overall aim was to develop a simple algorithm that only returns a set of line bounding boxes and does not incorporate additional features, such as general text direction (left-to-right, top-to-bottom, etc.) detection, for example. \\
\end{itemize}
Having laid out the context of this work, the remainder of this paper is structured in two sections, followed by concluding statements. The first one formally defines the algorithm, while reusing the same Figure~\ref{fig:input} example input to illustrate intermediate results. The subsequent section evaluates the method and draws the comparison to \textsc{Bench}.

\section{Algorithm}
The literature has shown that text line segmentation methods can essentially be classified in two categories (\citet{u-net}). First, well documented by \citet{survey}, there are image processing methods, such as connected-components, projections and smearing analysis. Second, there are learning-based methods, which are better suited for more complex layouts and heterogeneous documents (\citet{learning}).
\\ \\
Given the single-column BnL data, the interplay of morphology and histogram projections, which can lead to encouraging results (\citet{combi}), was retained for this work. The nature of combined techniques is also the reason why the proposed segmentation method is referred to as \textsc{CombiSeg}. In the following, different subsections will cover the algorithm's input requirements, the functioning of the segmentation itself and the resulting output.
\subsection{Input Assumption}
To define the scope of the approach and consequently clarify the input requirements, it is sufficient to state that the algorithm expects an input image fulfilling two properties: \\
\begin{enumerate}
    \item The input image, in the following denoted as $I^b$, must be presented in a binarized form, composed of ones (text/white) and zeros (background/black) only. This paper represents $I^b$ as a 2D array, with the height $h(I^b)$ seen as the first, and the width $w(I^b)$ as the second dimension. It follows that the highest indices are $h(I^b)\!-\!1$ and $w(I^b)\!-\!1$ respectively. \\
    To comply with this, BnL relies on Otsu's binarization method (\citet{otsu}) as implemented in the $OpenCV$ library (\citet{opencv}). \\
    \item \textsc{CombiSeg} is not designed to perform layout analysis (e.g. identify text blocks within a full newspaper page). That's why the second requirement is that $I^b$ is a block of text, solely containing horizontal and typed text lines (not tested for handwriting). In the example of the BnL use case, the employed dataset was already pre-segmented into blocks of text, mostly representing paragraphs or small articles.
\end{enumerate}
\begin{figure}[H]
\centering
\fbox{\includegraphics[width=0.66\textwidth]{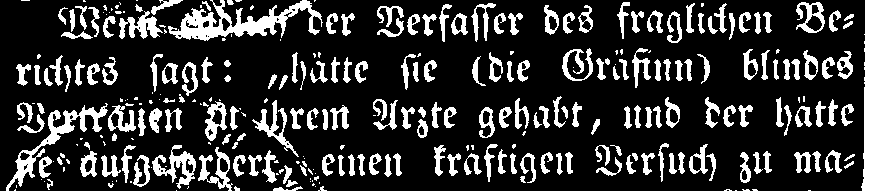}}
\captionsetup{format=hang, margin=0.0\textwidth}
\caption{Valid example \textsc{CombiSeg} input.}
\label{fig:inverted}
\end{figure}
\subsection{Segmentation}

Given the correct input, the actual algorithm can be be summed up as a sequence of three consecutive steps. \\
\begin{enumerate}
    \item $Morphology$ \\
    First, a set of morphological image operations is applied on $I^b$, in order to create a new processed version, denoted as $I^p$. \\
    \item $Components$ \\
    Next, $I^p$ is used to search for connected components, with each component potentially already representing a single text line. \\
    \item $Histogram$ \\
    Finally, a horizontal histogram projection is in turn created for each component, being the basis for possible additional text line splits. \\
\end{enumerate}
The algorithm is also subject to an ordered set of parameters, allowing slight tuning and referred to as $params\!=\!\{p_1, p_2, p_3, p_4, p_5, p_6, p_7, p_8\}$. \\ \\
Having laid out the high level plan for \mbox{\textsc{CombiSeg}}, this article will proceed with explanations regarding the precise working of the three steps involved.
\subsubsection{Morphology}

\begin{figure}[H]
\centering
\fbox{\includegraphics[width=0.66\textwidth]{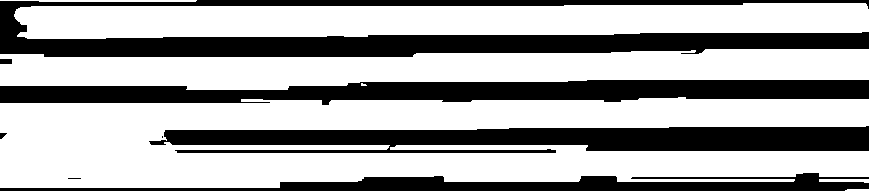}}
\captionsetup{format=hang, margin=0.0\textwidth}
\caption{Processed image $I^p$ originating from binary image $I^b$.}
\label{fig:final}
\end{figure}

To begin with, a set of sequential image operations is applied to obtain a processed image version, with the intention that every connected pixel component represents a text line. For explanation purposes, this paper will, in the following, make use of the common image operations $opening\,(\circ)$, $dilation\,(\oplus)$, $addition\,(+)$, $subtraction\,(-)$ and  $inversion\,(^{-1})$. To provide an example,
\begin{equation}
I^p\gets I^b\circ E_{w\times h}
\end{equation}
states that image $I^p$ is obtained by opening image $I^b$ using structuring element $E_{w\times h}$, which is filled with ones only, for width $w$ and height $h$. Image operations are also considered to return a modified copy, rather than changing the image that's being operated on.
\begin{listing} [H]
  \begin{lstlisting}[mathescape]
    $\textbf{morph}(I^b)$
        $I^p\gets I^b-\big((I^b\circ E_{1\times p_1}) + (I^b\circ E_{p_1\times 1})\big)$
        $I^p\gets(I^p\oplus E_{p_2\times 1})^{-1}$
        $S\gets \big(I^p - (I^p\circ E_{1\times p_3})\circ E_{p_4\times 1}\big)\oplus E_{p_5\times 1}$
        return $(I^p+S)^{-1}$
    \end{lstlisting}
\caption{Transforms binary $I^b$ into processed $I^p$, featuring connected components as potential text lines.}
\label{alg:alg1}
\end{listing}
Given the pseudocode seen in Algorithm~\ref{alg:alg1}, the line statements can be explained as follows: \\ \\
$\bullet$ Line 2: \\ \\
Considered to be a preprocessing operation. The first statement of Algorithm~\ref{alg:alg1} tries to remove long vertical and horizontal elements (borders, lines, frames, etc.) from $I^b$ through opening, followed by subtraction. Hence, every element has to have minimum height/width of $p_1$ to be considered. The main motivation here is to deal with the potentially cropped nature of the input images. For instance, in the concrete example of $I^b$ having been cropped out of a newspaper page, imprecise croppings and the inclusion of neighbouring article separator lines can be the cause for segmentation algorithms to falsely combine multiple successive text lines.
\begin{figure}[H]
\centering
\fbox{\includegraphics[width=0.66\textwidth]{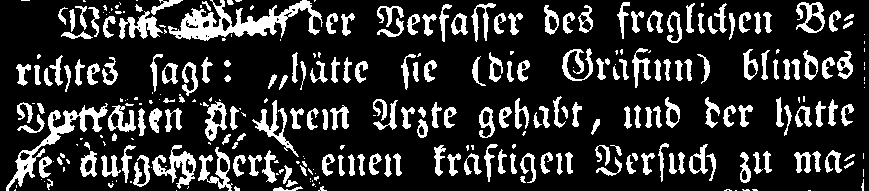}}
\captionsetup{format=hang, margin=0.0\textwidth}
\caption{Intermediate result after $morph$ preprocessing step, almost entirely removing the border line at the very right.}
\label{fig:preproc}
\end{figure}
\noindent $\bullet$ Line 3: \\ \\
It follows that all text areas in the preprocessed version $I^p$ are horizontally dilated using structuring element $E_{p_2\times 1}$, in an attempt to concatenate consecutive characters of the same text line (connected components creation). The obtained result is subsequently inverted (white background) to allow incoming background modifications.
\begin{figure}[H]
\centering
\fbox{\includegraphics[width=0.66\textwidth]{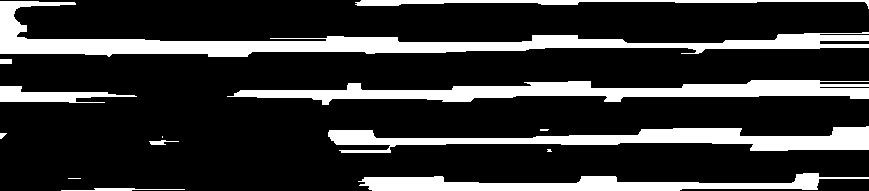}}
\captionsetup{format=hang, margin=0.0\textwidth}
\caption{Horizontally dilated and inverted version.}
\label{fig:dilatedinverted}
\end{figure}
\noindent $\bullet$ Line 4: \\ \\
The next statement starts by removing vertical elements from $I^p$ using an opening operation (protective step). This helps to prepare the background of the image (as seen in Figure~\ref{fig:dilatedinverted}), in a way that it should prevent the next step (horizontal dilation) to cause damage to the text areas. This next step consist of searching for horizontal background separators. Once found, they are dilated horizontally to increase their width. This should break unwanted connections between successive text lines (e.g. second and third text line in Figure~\ref{fig:dilatedinverted}), that might have been created by the very first dilation. The protective first step thus attempts to avoid that background separators are detected in front or after text lines, that don't extend over the entire image width, and consequently cut into the text areas. The result of the entire statement is stored in $S$, representing the separators between text line areas.
\begin{figure}[H]
\centering
\fbox{\includegraphics[width=0.66\textwidth]{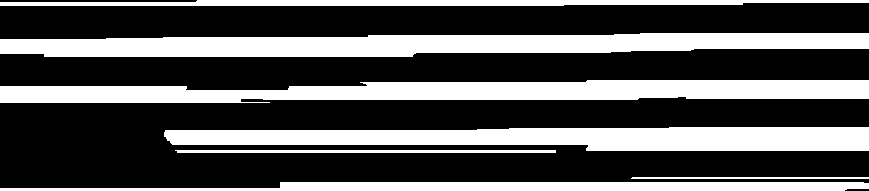}}
\captionsetup{format=hang, margin=0.0\textwidth}
\caption{Representation of text area separators $S$ that have been reinforced compared to Figure~\ref{fig:dilatedinverted}.}
\label{fig:backgroundilation}
\end{figure}
\noindent $\bullet$ Line 5: \\ \\
Adding $S$ to the inverted image version is equivalent to the subtraction from the regular one. This explains the process of the last algorithm statement. As mentioned in the previous paragraph, the separators attempt to break unwanted text area connections. The final $I^p$ of the example input image, as returned by Algorithm~\ref{alg:alg1}, can be seen with Figure~\ref{fig:final}. Comparing with Figure~\ref{fig:dilatedinverted}, the addition of $S$ ends up being the catalyst for the separation of text lines two and three.
\subsubsection{Components}

Using $I^p$, all connected white pixel components are identified. For this purpose, the 4-way-connectivity is used (i.e. diagonal pixels are not considered connections). Next, all component bounding boxes, using coordinates within $I^p$, are determined and stored in a set of no particular order. \\ \\
Let $n$ denote the number of found components. To determine bounding $box_i$ of the $i$th component, also let $X_i$ and $Y_i$ represent the set of all x-coordinates and y-coordinates in $I^p$, of all pixels belonging to the $i$th component. It is defined that
\begin{align}
    box_i\!=\!\big[min(X_i),\!min(Y_i),\!max(X_i),\!max(Y_i)\big].
\end{align}
Finally, the set $boxes$ is used to store a first version of the bounding boxes that the algorithm will ultimately return. At the same time, bounding boxes that don't have minimum height $p_6$ are discarded in a way that
\begin{align}
&boxes = \big\{box_i\;|\;box_i[3]\!-\!box_i[1]\geq p_6\big\},\;\text{with}\;1\leq i\leq n.
\end{align}
The edge case, where $boxes$ is bound to be an empty set, is covered through the assignment of
\begin{align}
&\text{if }|boxes|=0\text{:} \\ \nonumber
&\quad boxes = \big\{[0, 0, w(I^p)\!-\!1, h(I^p)\!-\!1]\big\}.
\end{align}
Using the process described in this subsection, a $comp$ function is defined as
\begin{align}
    comp(I^p) \rightarrow boxes.
\end{align}
\begin{figure}[H]
\centering
\fbox{\includegraphics[width=0.66\textwidth]{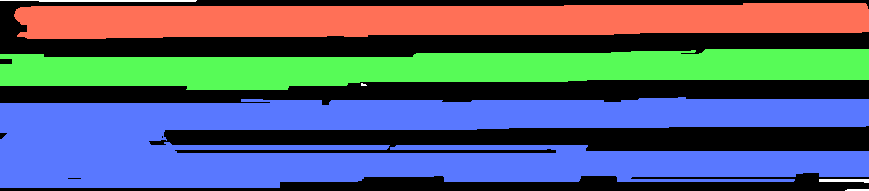}}
\captionsetup{format=hang, margin=0.0\textwidth}
\caption{Visualization of $n\!=\!3$ connected components with minimum height $p_6$.}
\label{fig:components}
\end{figure}
\subsubsection{Histogram}

The final but essential step of \textsc{CombiSeg} is the analysis of horizontal histogram projections, one created for every bounding box stored in $boxes$. The global projection of $I^b$ is denoted as $H^*$, an array storing the number of white pixels in every row of the image, s.t.
\begin{align}
&H^*[j] = \sum_{k=0}^{w(I^p)-1}I^b[j][k],\;\text{with}\;0\leq j\leq h(I^p)\!-\!1.
\end{align}
\begin{figure}[H]
\centering
\fbox{\includegraphics[width=0.66\textwidth]{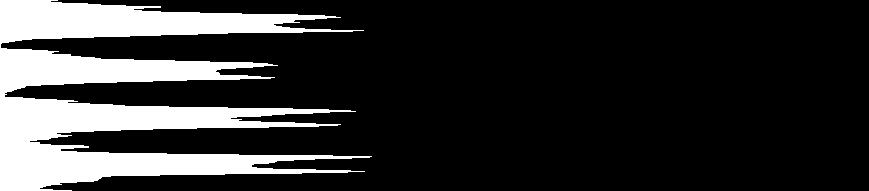}}
\captionsetup{format=hang, margin=0.0\textwidth}
\caption{Representation of global projection $H^*$ of example image.}
\label{fig:projection}
\end{figure}
\noindent Through the selection of the subset values $H^*\big[box_i[1]..box_i[3]\big]$ (i.e. all values that fall in the vertical component range) and their subsequent reordering, a sub-projection $H_{i}$ is created for every $i$th component from $1$ to $n$. This sub-projection is considered to be a dictionary, ordered by decreasing histogram values, while the y-coordinates are figuring as dictionary keys. This process is in the following referred to as
\begin{align}
    subproj(H^*, box_i)\rightarrow H_i.
\end{align}
The sub-projection is then passed to the $analysis$ procedure (Algorithm~\ref{alg:alg3.1}), a method proposed by \citet{usedAlgo}, that analyzes the locations of peaks in the histogram data. Next, based on the identified peaks, the procedure suggests division y-coordinates to split the underlying connected component bounding box. \\ \\
This is done by identifying peaks using a relative threshold $\alpha_t$, having a value proportional to the height of the highest location of the potential peak, which is equal to the height $h$ at $coord$. The set $C$ keeps track of already analyzed coordinates. Should the new range of peak coordinates $R$ be disjoined with $C$, a new peak has been identified and its delimiters $p_{start}$ and $p_{end}$ are added to the peaks set $P$.
\begin{listing} [H]
  \begin{lstlisting}[mathescape]
    $\textbf{analysis}(H_i, t\!=\!p_7)$
        $C\gets \emptyset$
        $P\gets \emptyset$
        for $coord$ in $H_i$
            $h\gets H_i[coord]$
            if $h< 0.1\!\times\!max(H_i)$ break
            if $coord\notin C$ 
                $\alpha_t\gets t\!\times\!h$
                $p_{start},\!p_{end}\!\gets\!bounds(H_i,\!coord,\!\alpha_t)$
                $R\gets [p_{start}..p_{end}]$
                if $C\cap R=\emptyset$
                    $P.append\gets p_{start}\text{, }p_{end}$
                $C\gets C\cup R$
        return $valleys(H_i,P)$
    \end{lstlisting}
\caption{Analyzes sub-projection and returns possible y divisor coordinates.}
\label{alg:alg3.1}
\end{listing}
\noindent The $analysis$ procedure in turn calls two different helper functions: \\
\begin{itemize}
    \item $bounds(H_i, coord, \alpha_t)$ \\ \\
    Accepts an ordered dictionary sub-projection $H_i$, a y-coordinate $coord$ and the height threshold $\alpha_t$ as input. The function computes and returns the delimiting coordinates $p_{start}$ and $p_{end}$ of the histogram peak located at $coord$. This is done by iteratively increasing/decreasing $coord$ by one, depending on the bound that should be discovered. A delimiting coordinate has been found, once the value of $H_i$, for the given new coordinate, falls below the relative threshold $\alpha_t$. It follows that
    \begin{align}
        H_i[p_{start}]\geq \alpha_t\;\land\;H_i[p_{start}\!-\!1]<\alpha_t\;\land\;H_i[p_{end}]\geq \alpha_t\;\land\;H_i[p_{end}\!+\!1]<\alpha_t
    \end{align}
    is satisfied.
    \begin{figure}[H]
    \centering
    \includegraphics[width=0.66\textwidth]{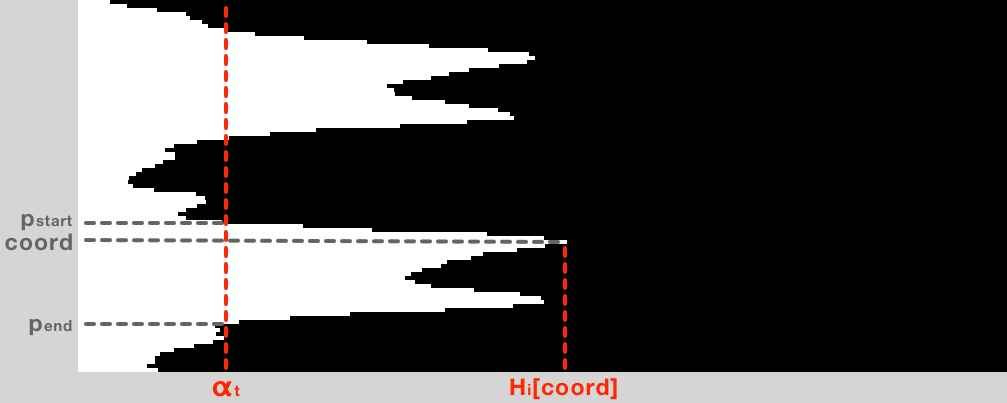}
    \captionsetup{format=hang, margin=0.0\textwidth}
    \caption{Visualization of $bounds$ applied to sub-projection of third component (text lines three and four).}
    \label{fig:subproj}
    \end{figure}
    \item $valleys(H_i, P)$ \\ \\
    The same sub-projection, together with all the computed peak delimiting coordinates $P$ is passed here. Once the set $P$ is ordered increasingly, the procedure computes splitting y-coordinates, representing valley locations in the histogram data. This is done by first discarding the first (lowest) and last (highest) value in $P$. Next, the range of coordinates between every consecutive pair of the remaining values ($[p_{end}..p_{start}]$ ranges) is considered. Finally, the coordinate of every range, that reveals the lowest histogram value in $H_i$, is appended to the returning $splits$ list. In case of $|P|<4$, no splitting coordinate is returned, since the projection features at most a single peak (i.e. connected component represents a single text line).\\
\end{itemize}
Now that it has been established how $splits$ can be computed for every connected component found in $I^p$, it remains to be defined how the splitting coordinates are to be used.
\begin{listing} [H]
  \begin{lstlisting}[mathescape]
    $\textbf{split}(splits, box_i)$
        $boxes^u\gets \emptyset$
        $splits\gets splits\cup box_i[3]$
        $last\gets box_i[1]$
        if $|splits|>0$
            for $s$ in $splits$
                if $s-last\geq p_6$
                    $box\gets[box_i[0], last, box_i[2], s]$
                    $boxes^u.append\gets box$
                $last\gets s$
        else
            $boxes^u.append\gets box_i$
        return $boxes^u$
    \end{lstlisting}
\caption{Potentially splits $box_i$ into multiple boxes using the $splits$ coordinates.}
\label{alg:splits}
\end{listing}
\noindent Algorithm~\ref{alg:splits} iterates through the $splits$ list and creates new bounding boxes by splitting at every coordinate $s$ in case the resulting box respects the minimum height of $p_6$. 
\begin{listing} [H]
  \begin{lstlisting}[mathescape]
    $\textbf{hist}(I^b, boxes)$
        $boxes^u\gets \emptyset$
        for $box_i$ in $boxes$
            $H_i\gets subproj(H, box_i)$
            $splits\gets analysis(H_i)$
            $boxes^u.append\gets split(splits, box_i)$
        return $boxes^u$
    \end{lstlisting}
\caption{Transforms initial $boxes$ into updated $boxes^u$ using histogram information of $I^b$.}
\label{alg:hist}
\end{listing}
\noindent Bringing everything together, an updated set $boxes^u$ is computed as indicated in Algorithm~\ref{alg:hist}.
\subsection{Output}

\begin{figure}[H]
\centering
\includegraphics[width=0.66\textwidth]{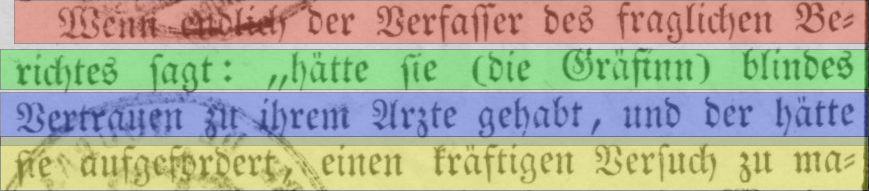}
\captionsetup{format=hang, margin=0.0\textwidth}
\caption{Visualization of the four resulting output bounding boxes of \mbox{\textsc{CombiSeg}}.}
\label{fig:result}
\end{figure}

Algorithm~\ref{alg:combigseg} ultimately provides the interactions between all three major steps involved in \textsc{CombiSeg}.
\begin{listing} [H]
  \begin{lstlisting}[mathescape]
    $\textbf{CombiSeg}(I^b)$
        $I^p\gets morph(I^b)$
        $boxes\gets comp(I^p)$
        $boxes^u\gets hist(I^b, boxes)$
        return $adjust(boxes^u)$
    \end{lstlisting}
\caption{The \textsc{CombiSeg} high level algorithm.}
\label{alg:combigseg}
\end{listing}
\noindent What remains to be clarified is the call to $adjust$ in Line~5. As a postprocessing procedure, the bounding boxes are modified one last time, in a way that: \\
\begin{itemize}
    \item They are sorted by increasing $box_i[1]$. \\
    \item Their height is increased using parameter $p_8$, s.t. updated $box_i$ equals
    \begin{align}
        box_i = \big[box_i[0], box_i[1]\!-\!p_8, box_i[2], box_i[3]\!+\!p_8\big].
    \end{align}
    This increase is justified by the fact that the background dilation in $morph$ can be the cause of a slight height reduction, which should be compensated in $adjust$. In addition, it can be that bounding boxes with a greater amount of padding are preferred. \\
    \item Lastly, $adjust$ merges all pairs of bounding boxes where $box_i$ completely contains $box_j$, in a way that
    \begin{align}
        (box_i[c]\leq box_j[c]\;\forall c\in [0..1])\;\land\; 
        (box_i[c]\geq box_j[c]\;\forall c\in [2..3]),\quad\text{with}\quad i\neq j.
    \end{align}
\end{itemize}
Optionally, $adjust$ provides the occasion to add custom rules defining the merging of vertically overlapping bounding boxes (see Subsection~\ref{section:post}). \\ \\
The result of the application of \textsc{CombiSeg} can be seen with Figure~\ref{fig:result}. In the case of the example input, the blue component in Figure~\ref{fig:components} was the basis for a successful detection (by $analysis$) of two different peaks, which have been $split$ and turned into individual bounding boxes (overcoming the potential background stamp issue).
\subsection{Parameters}

In order to adapt to a specific target dataset and its proprietary characteristics, the set of $params$  can be tuned accordingly. The respective parameters, together with their usage description, are listed below: \\
\begin{itemize}
    \item $p_1$: Preprocessing element height/width.
    \item $p_2$: First (text) dilation width.
    \item $p_3$: Background protection element height.
    \item $p_4$: Background separator width.
    \item $p_5$: Second (background) dilation width.
    \item $p_6$: Minimum text line height.
    \item $p_7$: Relative peak height threshold.
    \item $p_8$: Postprocessing height adjustment.
\end{itemize}
\subsubsection{Value Determination}

The parameter values, experimentally leading to the best results on a BnL test set (more details will follow in Section~\ref{section:results}), are given by
\begin{align}
 params=\{100, 90, 25, 35, 330, 14, 0.3, 5\}.
\end{align}
To provide some guidance on the approach of identifying (near) optimum values, the following three-step course of action is proposed: \\
\begin{enumerate}
    \item Largely independent of every other parameter in the set, are $p_1$, $p_6$ and $p_8$. Their values are generally determined by "preference choices". While $p_1$ should be considerably (depending on the aggressiveness of the desired preprocessing) greater than a regular text line height, $p_6$ should obviously be smaller to solely filter out noise. \\
    \item Next, there are the morphology related parameters $p_2$, $p_3$, $p_4$ and $p_5$. A similar order of magnitude, with respect to the ratio compared to the the average text line height (42.9 pixels for the BnL test set), should be chosen as a starting point in every case. Afterwards, a small amount of discrete values should be brute forced (e.g. deltas of $-10, 0$ and $+10$, leading to $3^4\!=\!81$ individual runs) by applying all combinations on a relatively small test set. It is of course essential to evaluate \textsc{CombiSeg} on the output of $morph$ and $comp$ only, discarding the subsequent $hist$ component. \\
    \item Lastly, $p_7$ should be determined independently for a given set of fixed morphology values. Hence, this final parameter can be learned in a supervised way through the overall performance of \textsc{CombiSeg}. \\
\end{enumerate}
\section{Experimental Results} \label{section:results}

For the purpose of evaluating the performance of \textsc{CombiSeg}, a BnL ground truth dataset is used and referred to as being a set of tuples $(I^b, gt)\!\in\!GT$. It contains 5,958 text block images cropped from historic newspapers with ground truth bounding boxes originating from human annotators. The number of text lines per sample ranges from 2 to 143 for a total of 114,625 lines. A \href{https://data.bnl.lu/wp-content/uploads/2021/09/bnl-ground-truth-newspapers-before-1878-raw.zip}{Raw Ground Truth Pack}, representing the copyright-free subset of $GT$, is available on the \href{https://data.bnl.lu/data/historical-newspapers/}{Open Data Platform} of BnL (OCR Datasets section).

\subsection{Evaluation Method} \label{section:evaluation}

In order to run \textsc{CombiSeg} against $GT$, an evaluation method is required.
Common comparison methods typically rely on the measures of recall/precision/F (\cite{measures}) or on the intersection over union (IoU) metric (\citet{iou}).
\\ \\
Since \textsc{CombiSeg} represents a rather simple case with perfectly horizontal and squared bounding information, which generally extend over the entire width, the chosen $loss$ function focuses on the number of unmatched text lines. Next to deducting the number of correctly $matched$ lines from cardinality $|gt|$ (recall), the function also punishes the detection of superfluous text lines (precision), while enforcing an overall upper bound of $|gt|$, in a way that
\begin{align}
loss(boxes^u, gt)\;=\;min\big(|gt|, |gt|\!-\!|matched|+max(0, |boxes^u|\!-\!|gt|)\big),\label{formula:loss}
\end{align}
where
\begin{align}
matched=\Big\{a\!\in\!gt\;\Big|\;\exists b\!\in\!boxes^u\;s.t.\; abs\Big(\frac{a[1]+a[3]}{2}\!-\!\frac{b[1]+b[3]}{2}\Big)\leq\theta\Big\}.\label{formula:matched} 
\end{align}
Commenting on (\ref{formula:matched}), a trivial definition that compares the middle y-coordinates of bounding boxes, is established. This way the $loss$ function represents the correct identification of individual text lines (main challenge in the historic dataset), rather than focusing on perfect bounding box dimensions (e.g. IoU). \\ \\
The threshold value $\theta$ is set to equal $14.3$, representing a third of the average bounding box height in $GT$ of $42.9$ pixels. Hence, a matched bounding box still has considerable vertical shifting possibilities, reinforcing the main objective, being text line identification. \\ \\
Using (\ref{formula:loss}), the accuracy of \textsc{CombiSeg} can be compared to \textsc{Bench}, by computing
\begin{align}
    acc(algo)=1-\frac{\sum_{(I^b, gt)\in GT}loss\big(algo(I^b), gt\big)}{\sum_{(I^b, gt)\in GT}|gt|}
\end{align}
for any algorithm $algo$.
\\ \\
Naturally, a shortcoming of $loss$ is that it does not account for partial text line croppings, resulting from inaccurate bounding box dimensions. However, no significant implication for the OCR accuracy is asserted for the BnL dataset. The conjecture is that the connected components part of \textsc{CombiSeg} generally helps to render rather precise bounding box dimensions. 
\subsection{Postprocessing} \label{section:post}
The conducted experiments required postprocessing to make both algorithms more comparable. This is backed by the fact that \textsc{Bench} tends to pursue a higher degree of fragmentation. Multiple bounding boxes are generally generated for text that is horizontally aligned correctly, but features particular large spaces between words. Since BnL considers those cases to represent one single text line, custom rules had to be introduced to merge bounding boxes based on their vertical positions. Those rules are equally applied to the output of both, \textsc{Bench} and \textsc{CombiSeg} (in the context of the call to $adjust$). \\ \\
Let $box_i[1]\!\leq\!box_j[1]$, with $i\!\neq\!j$, denote the starting y-coordinates of two successive bounding boxes in the list of all output boxes, ordered based on this value. A vertical overlap between $box_i$ and $box_j$ is defined as
\begin{align}
    o=max(0, box_i[3]-box_j[1]).
\end{align}
A merger to a combined single bounding box is performed, in case there is
\begin{align}
    \Big(\frac{o}{box_i[3]-box_i[1]} > 0.75\Big)\!\lor\!\Big(\frac{o}{box_j[3]-box_j[1]} > 0.75\Big)\!\lor\!\Big(\frac{o}{box_j[3]-box_i[1]} > 0.5\Big). \label{formula:merger}
\end{align}
Commenting on (\ref{formula:merger}), boxes are merged in case $o$ represents more than 75\% of the height of one of the single boxes, or more than 50\% of the combined height of both boxes.
\subsection{Results}

By applying the constructed evaluation method of Section~\ref{section:evaluation}, together with the postprocessing rules of Section~\ref{section:post}, the experimental results based on $GT$ are
\begin{align}
    acc(\textsc{CombiSeg})&>acc(\textsc{Bench}) \nonumber \\
    \implies1-\frac{890}{114,625}&>1-\frac{2066}{114,625} \nonumber \\
    \implies0.992&>0.982.
\end{align}
Although observing improved accuracy is welcoming, the main objective of BnL was to improve computational efficiency with \textsc{CombiSeg}. Running both algorithms against $GT$ on the exact same hardware, delivered very promising results related to the average processing time, in a way that
\begin{align}
    pt(\textsc{CombiSeg}) &< pt(\textsc{Bench}) \nonumber \\
    \implies\frac{101,778}{5,958}&<\frac{4,583,208}{5,958} \nonumber \\
    \implies17.08&<769.25,
\end{align}
using
\begin{align}
    pt(algo)=\frac{total\;processing\;time\;(ms)}{number\;of\;processed\;images}.
\end{align}
The implementation of \textsc{CombiSeg} is based on methods from the $OpenCV$ library (\citet{opencv}) and can be accessed as part of the OCR project, which is publicly \href{https://github.com/natliblux/nautilusocr}{available}. On the other hand, the implementation of \textsc{Bench} was run as published with version $2.0.8$ of the software.
\section{Conclusion}

In summary, the proposed segmentation method \textsc{CombiSeg} is a perfect fit for the BnL use case of segmenting newspaper articles into individual text lines. The algorithm performs well in terms of both, accuracy and speed, thus having a positive effect on the overall OCR initiative. On top of that, it has the advantage of not requiring a separate training phase, as it is the case for model-based machine learning approaches. Its overall robustness originates from the combination of two techniques, commonly used in the literature. Morphological operations and horizontal histogram projections are combined, in a way that they complete each other and cancel out their weaknesses. \\
\begin{itemize}
    \item Generating connected components through dilation has the advantage of obtaining rather precise coordinate information. However, this technique is prone to be fooled by background noise or closely adjacent lines, leading to unwanted connections and potentially major segmentation mistakes. \\
    \item Histogram projections, on the other hand, are less sensitive to anomalies and mostly overcome them by considering the integrity of image rows. Nonetheless, evaluating the histogram data and finding the correct bounding box through peak/valley detection, is not a trivial job. This is where the preliminary morphology work is able to help out. \\
\end{itemize}
Analyzing the limitations of \textsc{CombiSeg}, it should be stated that input images with a strong degree of skew or rotation can be the cause for some inaccuracies, such as the partial inclusion of a successive second text line in the same bounding box. Given that the algorithm is designed to strictly return horizontal boxes, a future BnL road-map could involve the expansion of the preceding binarization step, to include more advanced image processing, such as deskewing, rotation and image cleaning.
\\ \\
Setting aside this reflection on future work, \textsc{CombigSeg} marks a promising segmentation algorithm candidate, especially for OCR purposes in relation with historic datasets that are in a binarized, single-column format.
\bibliographystyle{plainnat}
\bibliography{references}
\end{document}